\pdfoutput=1

\documentclass[11pt]{article}

\usepackage{xcolor}
\definecolor{gold}{RGB}{255,192,0}
\definecolor{green}{RGB}{112,173,71}

\usepackage[linesnumbered,ruled,vlined]{algorithm2e}
\let\oldnl\nl
\newcommand{\nonl}{\renewcommand{\nl}{\let\nl\oldnl}}

\usepackage[preprint]{acl}

\usepackage{times}
\usepackage{latexsym}

\usepackage{amsmath}
\usepackage{amssymb}

\usepackage[T1]{fontenc}

\usepackage[utf8]{inputenc}

\usepackage{microtype}

\usepackage{inconsolata}

\usepackage{graphicx}

\title{Efficient Code LLM Training via Distribution-Consistent and Diversity-Aware Data Selection}

\author{Weijie Lyu \and Sheng-Jun Huang \\
  Nanjing University of \\
  Aeronautics and Astronautics \\
  \texttt{\{lvweijie, huangsj\}@nuaa.edu.cn} \\\And
  Xuan Xia \\
  Shenzhen Institute of Artificial \\
  Intelligence and Robotics for Society \\
  \texttt{xiaxuan@cuhk.edu.cn} \\}

\begin{document}
\maketitle

\begin{abstract}
Recent advancements in large language models (LLMs) have significantly improved code generation and program comprehension, accelerating the evolution of software engineering. Current methods primarily enhance model performance by leveraging vast amounts of data, focusing on data quantity while often overlooking data quality, thereby reducing training efficiency. To address this, we introduce an approach that utilizes a parametric model for code data selection, aimed at improving both training efficiency and model performance. Our method optimizes the parametric model to ensure distribution consistency and diversity within the selected subset, guaranteeing high-quality data. Experimental results demonstrate that using only 10K samples, our method achieves gains of 2.4\% (HumanEval) and 2.3\% (MBPP) over 92K full-sampled baseline, outperforming other sampling approaches in both performance and efficiency. This underscores that our method effectively boosts model performance while significantly reducing computational costs. Code is available at 
\href{https://github.com/Kyle-Lyu/efficode-finetune}{here}.
\end{abstract}


\section{Introduction}

Recent years have witnessed remarkable progress in large language models (LLMs), particularly in code-related domains \cite{hurst2024gpt, liu2024deepseek, guo2025deepseek}. Open-source code models \cite{li2023starcoder, lozhkov2024starcoder, guo2024deepseek} have significantly advanced academic research by demonstrating strong capabilities in code generation and program comprehension. Through large-scale pre-training, these models provide intelligent support across multiple programming languages and development environments, accelerating the evolution of software engineering intelligence. Meanwhile, instruction tuning has proven to be an effective method to enhance model performance \cite{wei2021finetuned, chung2024scaling}. By fine-tuning on large-scale instruction data, models better align with human intent and excel in specific tasks. However, high-quality human-annotated data is scarce and costly to obtain. To address this, researchers have proposed various methods for generating data to expand instruction datasets \cite{wang2023self, gunasekar2023textbooks}.

Recent studies \cite{zhou2023lima, xia2024less} emphasize that data quality is more important than quantity, highlighting the significance of representative data selection for training efficiency. To address this, many approaches \cite{chen2024alpagasus, lu2024instag, wang2024inscl} leverage advanced LLMs (e.g., ChatGPT) for data selection and annotation, thereby optimizing data quality and diversity. However, these methods face economic challenges when scaling to large datasets, and most algorithms target general tasks rather than code-specific data selection.

Inspired by ActiveFT \cite{xie2023active}, we introduce the parametric model into code data selection to improve training efficiency and model performance. Unlike traditional discrete selection methods, we operate in feature space for more effective data curation. Specifically, we first map each data sample to a high-dimensional feature space using a feature encoder. Then, we construct the parametric model based on these feature representations. Our goal is to ensure that the feature distribution of the selected subset closely matches the original dataset’s while maximizing the diversity within the subset. To achieve this, we continuously optimize the parametric model through a loss function that balances distribution consistency and diversity constraints. Finally, based on the optimized parametric model, we select samples most similar to the parameters to construct a high-quality subset.

We conducted extensive experiments to validate the effectiveness of our method. We mixed multiple datasets to construct a training set containing 92K Python samples and trained the model based on DeepSeek-Coder-Base-6.7B. The results show that, with only 10K sampled data, our method achieved 69.5\% on HumanEval and 77.2\% on MBPP, surpassing full-data training by 2.4\% and 2.3\%, respectively. Additionally, our method outperforms other sampling approaches across various data scales while requiring minimal sampling time. This demonstrates that our approach not only effectively selects high-quality data but also significantly improves model performance and computational efficiency.

The contributions of our work are summarized as follows:

\begin{itemize}
    \item As far as we know, we are the first to introduce the parametric model into code data selection. By ensuring distribution consistency and diversity, we successfully identify high-quality data, significantly improving model performance.

    \item We perform data selection in the feature space, avoiding traditional discrete selection methods. This greatly enhances sampling and training efficiency.

    \item Extensive experiments validate the effectiveness of our method. The results show that using only 10K sampled data outperforms full-data training. Moreover, our method surpasses other sampling methods in both performance and sampling time across various data scales.
\end{itemize}

\section{Related Work}

\subsection{Code Large Language Models}

The emergence of large language models (LLMs) has significantly advanced the intelligence-driven transformation of software engineering, particularly demonstrating breakthrough capabilities in code generation and program comprehension tasks. The open-source community has developed multiple high-performance code LLMs, with notable representatives including CodeLlama \cite{roziere2023code}, DeepSeek-Coder \cite{guo2024deepseek}, CodeGemma \cite{team2024codegemma}, and Qwen2.5-Coder \cite{hui2024qwen2}. 
CodeLlama, built upon the Llama 2 \cite{touvron2023llama} architecture through continued pre-training on 500B code tokens, achieves performance comparable to commercial models in standardized benchmarks.
The DeepSeek-Coder series, trained from scratch on 2T high-quality multilingual code corpora, demonstrates significant superiority over the same-scale counterparts in benchmark evaluations.
The Qwen2.5-Coder series are built upon the Qwen2.5 \cite{yang2024qwen2} architecture and continue training on a vast corpus of over 5.5 trillion tokens, surpassing closed-source models like GPT-4 \cite{achiam2023gpt} across multiple benchmarks and establishing new state-of-the-art records.
These open-source models, through architectural innovations and training strategy optimizations, not only deliver highly customizable code intelligence solutions but also construct efficient technical infrastructure for both academic research and industrial applications.

\subsection{Instruction Fine-tuning}

Instruction fine-tuning has been demonstrated as a crucial approach to enhance model performance and align models with human preferences. The study \cite{chung2024scaling} indicates that scaling both the number of tasks and the model size through instruction fine-tuning yields performance improvements across various model classes.
For instance, WizardCoder \cite{luowizardcoder} leverages the Evol-Instruct \cite{xu2024wizardlm} method to iteratively evolve the complexity of the CodeAlpaca \cite{codealpaca} dataset. This approach results in a fine-tuning dataset consisting of approximately 78K highly complex programming instructions, thereby improving the performance of CodeLLama-Python-34B to 73.2\% on the HumanEval benchmark.
Similarly, Magicoder \cite{wei2024magicoder} proposes the OSS-Instruct approach, which generates highly diverse instruction data by using open-source code snippets. This approach successfully generated a dataset containing approximately 75K entries.
Additionally, WaveCoder \cite{yu2024wavecoder} makes full use of open-source code data through a carefully designed generator-discriminator data synthesis framework to generate high quality and diverse instruction data in multi-task scenarios.

\subsection{Data Selection for Efficient Training}

Instruction fine-tuning typically requires large amounts of data, but research such as LIMA \cite{zhou2023lima} has pointed out that data quality is more crucial than quantity. Therefore, selecting the most valuable data to improve training efficiency has become a focal point of research.
DEITA \cite{liu2024makes} focuses on the complexity, quality and diversity of the data, designing a multifaceted approach to select instruction data. 
Based on the Evol-Instruct technique, ChatGPT is used to augment the instructions, and the instructions are then evaluated for complexity and quality by specially trained scorers.
Quantifiable metrics like PPL \cite{ankner2024perplexed}, IFD \cite{li2024quantity}, and Superfiltering \cite{li2024superfiltering} focus on identifying hard samples that are difficult to learn. 
DQ \cite{zhou2023dataset} integrates data distillation and coreset selection \cite{iyer2021submodular} techniques, emphasizing the selection of diverse data.
Some methods \cite{chen2024alpagasus, lu2024instag, xu2023rethinking} that rely on external oracles use powerful language models, such as ChatGPT, for data selection. However, due to cost constraints, utilizing external oracles is not always feasible.

Overall, many data selection algorithms are primarily designed for general tasks, and the integration of parametric models into code data selection remains underexplored.

\section{Methodology}

\begin{figure}[t]
  \includegraphics[width=1.0\linewidth]{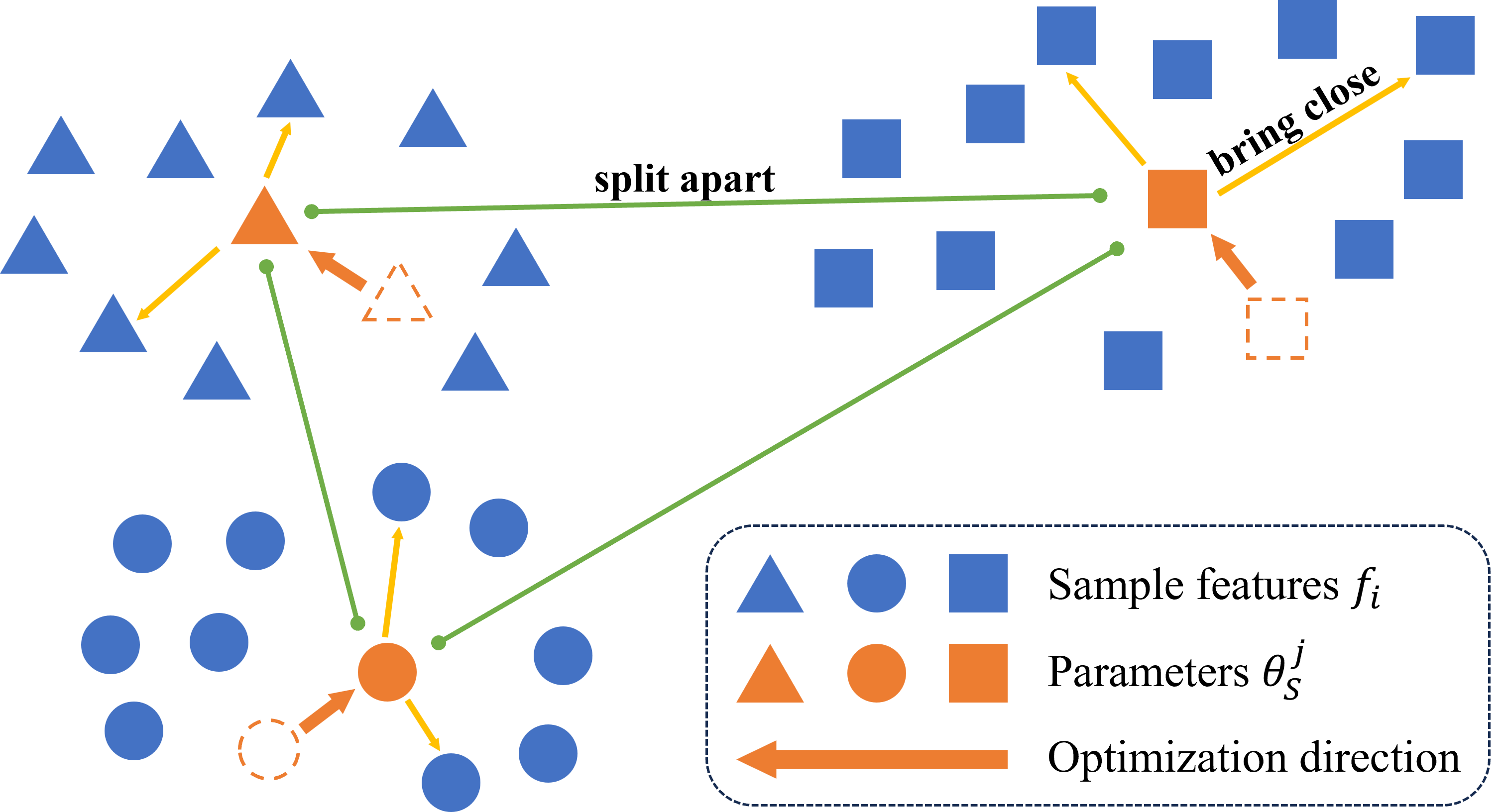}
  \caption {\textbf{Parametric Model Optimization Process:} By optimizing the loss in Equation~\ref{eq:7}, each parameter $\theta_S^j$ is attracted to nearby sample features(\textcolor{gold}{gold} in the figure, Equation~\ref{eq:5}) and repelled by other parameters $\theta_S^k, k \neq j$ (\textcolor{green}{green} in the figure, Equation~\ref{eq:6}).}
  \label{fig:optimization}
\end{figure}

Our methodology aims to identify and select high-quality, representative data samples so that training on the curated subset yields better performance than training on the entire dataset. 
Inspired by ActiveFT \cite{xie2023active}, we incorporate the parametric model into the code data selection process to improve both training efficiency and model performance. 
We start by defining the data selection task in Section~\ref{sec:task_definition}. 
Then we introduce the integration of parametric models into data selection in Section~\ref{sec:data_selection}, as illustrated in Figure~\ref{fig:optimization}. 
Finally, the implementation details are provided in Section~\ref{sec:implementation_details}.

\subsection{Task Definition} 
\label{sec:task_definition}

Given a large instruction tuning dataset $D = \{x_1, x_2, \ldots, x_n\}$, where each $x_i = (I_i, C_i)$ represents an individual instruction-code pair, our goal is to select a subset $S_\pi^{m} \subset D$ of size $m$ using a selection strategy $\pi$. 
The performance of the model after fine-tuning on $S_\pi^{m}$ is denoted by $P(S_\pi^{m})$ and is used to evaluate the effectiveness of the selected subset. The optimal strategy $\pi^*$ under a fixed budget $m$ is defined as:

\begin{equation}
    \pi^* = \arg\max_\pi P(S_\pi^{m}).
    \label{eq:1}
\end{equation}

\subsection{Data Selection with Parametric Model}
\label{sec:data_selection}

During data selection, we consider two key factors: On one hand, the selected subset $S$ should have the distribution as close as possible to that of the original dataset $D$. On the other hand, the subset $S$ should maintain the diversity. By balancing these two aspects, we aim to select a representative subset that covers edge cases in the original dataset, thereby enhancing model performance.

Compared to selecting in the discrete data space,  it is more efficient and feasible to conduct data selection in the feature space. We map each data sample $x_i$ to the high-dimensional feature space using a feature encoder $E$, resulting in $f_i = E(x_i)$ which is then normalized. After normalization, we obtain the feature set of the dataset $D$ as $F_D = \{f_i\}_{i \in [n]}$, whose distribution is denoted by $p_{F_D}$.

Similarly, for the selected subset $S$, we define its feature set as $F_S = \{f_j\}_{j \in [m]}$, with its corresponding distribution $p_{F_S}$. Our objective is to find the optimal selection strategy $\pi^*$ as follows:

\begin{equation}
    \pi^* = \arg\min_\pi \left[M(p_{F_D},p_{F_S}) - \lambda \cdot R(S_\pi^m)\right],
    \label{eq:2}
\end{equation}
where $M(\cdot, \cdot)$ measures the distance between two feature distributions, $R(\cdot)$ evaluates the diversity of the selected subset, and $\lambda$ is a scaling factor that balances two terms. The first term in Equation~\ref{eq:2} aims to align the distributions, while the second term ensures the diversity within the subset.

It's challenging to directly optimize the selection strategy $\pi$ in the discrete space, so we adopt a parametric model $p_{\theta_S}$ to approximate $p_{F_S}$, where $\theta_S = \{\theta_S^j\}_{j \in [m]}$ are the continuous parameters. Each optimized parameter $\theta_S^j$ corresponds to the feature of a selected sample $f_j$, and we select the feature $f_j$ that is closest to $\theta_S^j$. Thus, the optimization objective is written as follows:

\begin{equation}
    \small \theta_{S,\pi^*} = \arg\min_{\theta_S} \left[M(p_{F_D},p_{\theta_S}) - \lambda \cdot R(\theta_S)\right] \text{
s.t. } ||\theta_S^j||_2 = 1.
    \label{eq:3}
\end{equation}

The key distinction between the features $F_S = \{f_j\}$ and the parameters $\theta_S = \{\theta_S^j\}$ is that $f_j$ is a discrete feature corresponding to a data sample, whereas $\theta_S^j$ is continuous in the feature space. By optimizing $\theta_S$, we search for ideal samples in the feature space that cover the original data distribution while maintaining dispersion.

The first term represents the distribution matching term $M(p_{F_D}, p_{\theta_S})$, which measures the similarity between the feature distribution of the dataset and the parametric model. It is defined as:

\begin{equation}
    c_i = \arg\max_{j\in[m]} sim(f_i,\theta_S^j), 
    \label{eq:4}
\end{equation}

\begin{equation}
    M(p_{F_D}, p_{\theta_S}) = -\mathop{E}\limits_{f_i \in F_D} \left[sim(f_i, \theta_S^{c_i}) / \tau \right],
    \label{eq:5}
\end{equation}
where $sim(\cdot, \cdot)$ is a similarity function, such as cosine similarity, and $\tau$ is a temperature parameter that controls the sensitivity of similarity measure. This term encourages the parameters to be well-spread across the feature space, effectively covering the distribution of the original features.

The second term is the diversity regularization $R(\theta_S)$, which promotes diversity within the selected subset by minimizing the similarity between the selected samples. It is defined as:

\begin{equation}
    \small R(\theta_S) = -\mathop{E}\limits_{j \in [m]} \left[ \log \sum_{k \neq j, k \in [m]} \exp \left( sim(\theta_S^j, \theta_S^k) / \tau \right) \right].
    \label{eq:6}
\end{equation}

We jointly optimize the following loss function to achieve the goal in Equation~\ref{eq:3}, where $\lambda$ is set to 1 by default.

\begin{equation}
    \begin{split}
        L = & M(p_{F_D}, p_{\theta_S}) - \lambda \cdot R(\theta_S) \\
        = & -\mathop{E}\limits_{f_i \in F_D} \left[\text{sim}(f_i, \theta_S^{c_i}) / \tau \right] \\
        & + \mathop{E}\limits_{j \in [m]} \left[ \log \sum_{k \neq j, k \in [m]} \exp \left( \text{sim}(\theta_S^j, \theta_S^k) / \tau \right) \right].
    \end{split}
    \label{eq:7}
\end{equation}

We optimize the loss function in Equation~\ref{eq:7} using gradient descent. Upon completion of the optimization, we find the features $\{f_j\}_{j \in [m]}$ that exhibit the highest similarity to $\theta_S^j$.

\begin{equation}
    f_j = \arg\max_{f_i \in F_D} \text{sim}(f_i, \theta_S^j).
    \label{eq:8}
\end{equation}

Finally, we collect the data samples corresponding to these selected features to form the instruction subset $S_\pi^m$, which will be used for fine-tuning.

\begin{algorithm}
\caption{Pseudo-code for Our Data Selection Algorithm}
\fontsize{10}{12}\selectfont
\label{alg:data_selection}

\KwIn{Dataset $D = \{x_i\}_{i \in [n]}$, feature encoder $E$, selection budget $m$, iteration number $T$ for optimization}
\KwOut{Subset $S = \{x_j\}_{j \in [m]}$}

\For{$i \in [n]$}{
    $f_i = E(x_i)$\;
}
\nonl \textcolor{blue}{\tcp{Construct $F_D = \{f_i\}_{i \in [n]}$ based on $D$, normalized to $||f_i||_2 = 1$}}

Randomly sample $\{f_j^0\}_{j \in [m]}$ from $F_D$, and initialize $\theta_S^j = f_j^0$\;

\nonl \textcolor{blue}{\tcp{Initialize the parameters $\theta_S = \{ \theta_S^j \}_{j \in [m]}$}}

\For{$iter \in [T]$}{
    Calculate the similarity between $\{f_i\}_{i \in [n]}$ and $\{\theta_S^j\}_{j \in [m]}$: $Sim_{i,j} = f_i^\top \theta_S^j / \tau$\; 
    $MaxSim_i = \max_{j \in [m]} Sim_{i,j} = Sim_{i,c_i}$\;
    
    \nonl \textcolor{blue}{\tcp{The Top-1 similarity between $f_i$ and $\theta_S^j$, $j \in [m]$}}
    
    Calculate the similarity between $\theta_S^j$ and $\theta_S^k$, $k \neq j$ for regularization: $RegSim_{j,k} = \exp({\theta_S^j}^\top \theta_S^k / \tau), k \neq j$\;
    
    $Loss = - \frac{1}{n} \sum_{i \in [n]} MaxSim_i + \frac{1}{m} \sum_{j \in [m]} \log \left( \sum_{k \neq j, k \in [m]} RegSim_{j,k} \right)$\;
    
    \nonl \textcolor{blue}{\tcp{Calculate the loss function in Equation~\ref{eq:7}}}
    
    $\theta_S = \theta_S - lr \cdot \nabla_{\theta_S} Loss$\;
    \nonl \textcolor{blue}{\tcp{Optimize the parameter through gradient descent}}
    
    $\theta_S^j = \theta_S^j / ||\theta_S^j||_2, j \in [m]$\;
    
    \nonl \textcolor{blue}{\tcp{Normalize the parameters to ensure $||\theta_S^j||_2 = 1$}}
}

\For{$j \in [m]$}{
    Find $f_j$ closest to $\theta_S^j$: $f_j = \arg \max_{k \in [n]} f_k^\top \theta_S^j$\;
    Find $x_j$ corresponding to $f_j$\;
}

Return the subset $S = \{x_j\}_{j \in [m]}$\;

\end{algorithm}

\subsection{Implementation Details}
\label{sec:implementation_details}

Algorithm~\ref{alg:data_selection} illustrates the implementation details of the data selection process. We use the pre-trained model \textit{all-mpnet-base-v2}\footnote{https://huggingface.co/sentence-transformers/all-mpnet-base-v2} from the sentence-transformers \cite{reimers-2019-sentence-bert} library as the feature encoder. This model has been trained on over 1 billion text pairs, effectively capturing the semantic information in instruction texts. For each data sample $x_i = (I_i, C_i)$, we encode the instruction text $I_i$ to obtain the feature vector $f_i = E(I_i) \in \mathbb{R}^{768}$, as the instruction text fully defines the task semantics. All features are then processed using L2 normalization, forming the normalized feature set $F_D = \{f_i\}_{i \in [n]}$.

Before optimizing the parametric model, the parameters $\theta_S$ are initialized by randomly sampling from the feature set $F_D$. During each iteration, to avoid memory overflow, we calculate the similarity between sample features and parameters in a batch-wise manner. Subsequently, we update $c_i$ according to Equation~\ref{eq:4}. Finally, we compute the loss function in Equation~\ref{eq:7} and update the parameters $\theta_S$ using gradient descent.

When the optimization process is finished, we find the sample feature $f_j$ that exhibit the highest similarity to $\theta_S^j$ according to Equation~\ref{eq:8}. The corresponding original sample $x_j$ is selected for subsequent fine-tuning.
This process ensures that the selected subset is highly representative and covers the edge cases in the original dataset, ultimately providing high-quality data support for the fine-tuning process.

\section{Experimental Setup}

\subsection{Dataset}

We use three open-source instruction datasets, including \emph{Evol-Instruct-Python-26K}\footnote{https://huggingface.co/datasets/mlabonne/Evol-Instruct-Python-26k}, \emph{CodeExercise-Python-27K}\footnote{https://huggingface.co/datasets/codefuse-ai/CodeExercise-Python-27k}, and \emph{OSS-Instruct-75K}\footnote{https://huggingface.co/datasets/ise-uiuc/Magicoder-OSS-Instruct-75K}.
The \emph{Evol-Instruct-Python-26K} dataset is the Python subset of the \emph{Evol-Instruct-80K}\footnote{https://huggingface.co/datasets/nickrosh/Evol-Instruct-Code-80k-v1} dataset, and its construction follows an iterative evolution strategy. Based on the CodeAlpaca \cite{codealpaca} instruction dataset, multi-round complexity enhancement operations are applied to the original problems using ChatGPT.
The \emph{CodeExercise-Python-27K} dataset is generated using the Camel \cite{li2023camel} framework, covering hundreds of Python-related topics including basic syntax and data structures, algorithm applications, database queries, machine learning, and more.
The \emph{OSS-Instruct-75K} dataset utilizes ChatGPT to generate programming problems and their corresponding solutions. Its uniqueness lies in the use of open-source code snippets as guidance for generation.
To maintain consistency in the programming language, we filter this dataset and only retain Python-related entries.
Finally, we merge these three datasets to obtain a mixed dataset, \emph{Mix-Python-92K}. We use this mixed dataset for training.

\subsection{Evaluation Benchmarks}

\textbf{HumanEval/HumanEval+.} 
HumanEval \cite{chen2021evaluating} comprises 164 Python programming problems designed to assess the ability of code generation models. Each problem is accompanied by approximately 9.6 test cases to check whether the generated code works as expected. HumanEval has become one of the most widely used benchmarks for evaluating the performance of code LLMs, making it a key tool in the field of artificial intelligence for coding. 
To enhance the rigor of the evaluation, HumanEval+ \cite{liu2023your} builds upon the original dataset by significantly increasing the number of test cases through the use of LLMs and mutation strategies, resulting in a more comprehensive evaluation benchmark.

\textbf{MBPP/MBPP+.} 
MBPP \cite{austin2021program} consists of approximately 1,000 Python programming challenges sourced from a crowd of contributors, targeting beginners in programming and focusing on core principles and the usage of the standard library. Each challenge includes a description, a solution and three tests to verify the accuracy.
To improve the reliability of the benchmark, MBPP+ \cite{liu2023your} extends the original dataset by incorporating a subset of hand-verified problems from the MBPP-sanitized dataset, ensuring that the tasks are well-defined and unambiguous. This enhances the benchmark’s reliability and suitability for more rigorous evaluations.

\begin{table*}[t]
    \centering
    \begin{tabular}{lcccccc}
    \hline
    Methods & Data Size & Sampling Time & HumanEval & HumanEval+ & MBPP & MBPP+ \\
    \hline
    Random & 10K & - & 64.6\% & 61.0\% & 74.3\% & 61.9\% \\
    DQ & 10K & 19.9h & 64.6\% & 60.4\% & 75.9\% & 62.4\% \\
    DEITA & 10K & 7.2h & 65.2\% & 61.0\% & 75.4\% & 63.0\% \\
    PPL & 10K & 3.6h & 62.2\% & 54.9\% & 74.6\% & 61.1\% \\
    IFD & 10K & 3.6h & 63.4\% & 57.3\% & 62.4\% & 46.6\% \\
    K-Center & 10K & 42.5 min & 64.6\% & 61.0\% & 74.9\% & 61.6\% \\
    Ours & 10K & \textbf{13.5min} & \textbf{69.5\%} & \textbf{63.4\%} & \textbf{77.2\%} & \textbf{63.2\%} \\
    \hline
    \end{tabular}
    \caption{Comparison of different sampling methods. All methods select 10K samples to train DeepSeekCoder-Base-6.7B model. Our method outperforms the others in terms of Pass@1 (\%) metric, while reducing sampling time.}
     \label{tab:sampling_methods}
\end{table*}

\subsection{Evaluation Metrics}

\textbf{Pass@k.} We use the Pass@k metric \cite{chen2021evaluating} to enhance the reliability of our evaluation process. We count the total number of generated samples that successfully passing all test cases, denoted as $k$, to compute the Pass@k.

\begin{equation}
    \text{Pass@}k: = \mathbb{E} \left[ 1 - \frac{\binom{n-c}{k}}{\binom{n}{k}} \right],
\end{equation}
where $n$ is the total number of generated samples for each problem and $c$ is the number of correct generated code samples passing all test cases($n > k \geq c$). In subsequent experiments, we compute the Pass@1 (\%) metric using greedy decoding.

\subsection{Training Details}

In our experiments, we use DeepSeekCoder-Base-6.7B as the base model and conduct training on eigth NVIDIA A800-80GB GPUs using PyTorch's Fully Sharded Data Parallel (FSDP) module for 3 epochs. Specifically, we employ the AdamW optimizer with a learning rate of 5e-5, a cosine learning rate scheduler, and 100 warmup steps. The maximum sequence length per batch is set to 4096 tokens with a global batch size of 512. To improve training efficiency, we utilize the Dynamic Pack \cite{lv2025data}. Model evaluation is conducted using the EvalPlus \cite{liu2023your} library.

For optimizing the parametric model, we adopt the same hyperparameter settings as in ActiveFT \cite{xie2023active}. Specifically, we use the Adam optimizer with 300 optimization iterations, a learning rate of 0.001, and set the temperature parameter $\tau$ in Equation~\ref{eq:7} to 0.07.

\begin{figure*}[t]
  \includegraphics[width=1.0\linewidth]{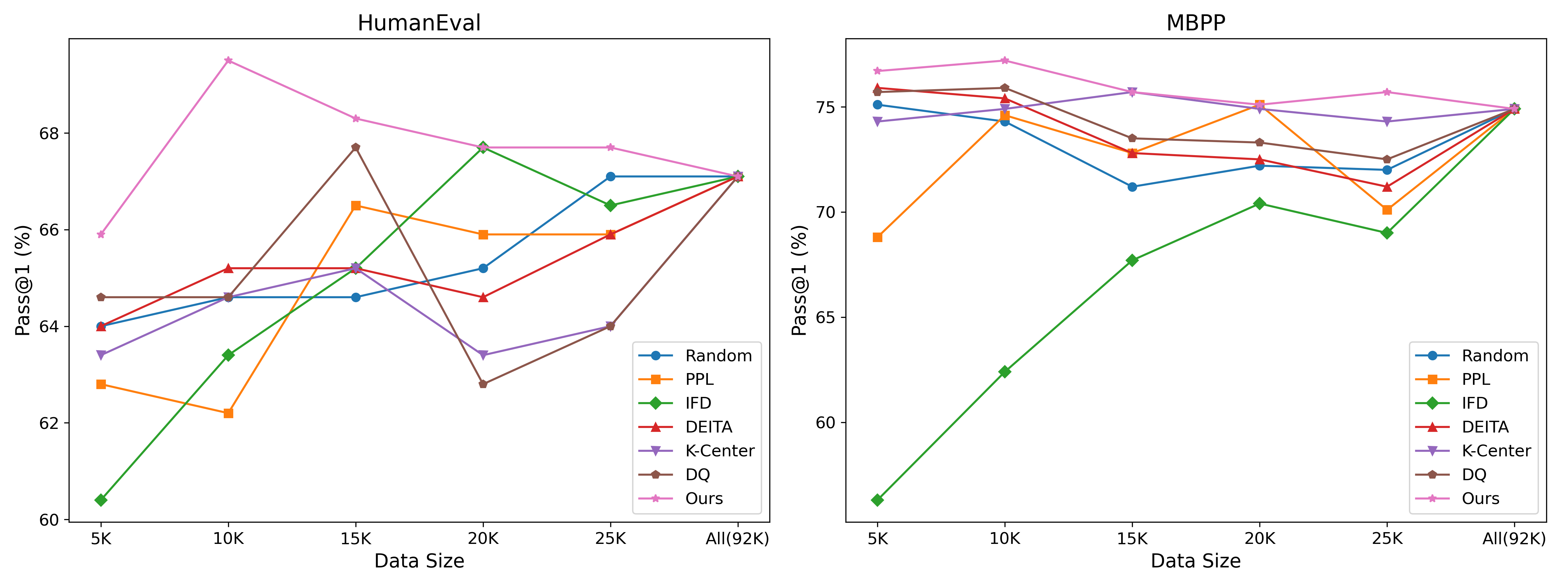}
  \caption {Performance comparison of different sampling methods with varying sampling quantities. Our method outperforms the other methods across different sample sizes, achieving the best performance with 10K samples.}
  \label{fig:sampling_quantity}
\end{figure*}

\section{Results}

\subsection{Sampling Methods}

We compared various sampling methods, with the experimental results presented in Table~\ref{tab:sampling_methods}. when sampling 10K samples, our method outperforms all baseline approaches across all benchmarks. Specifically, our method achieves 69.5\% on HumanEval, 63.4\% on HumanEval+, 77.2\% on MBPP, and 63.2\% on MBPP+.
These results demonstrate the importance of considering both distribution consistency and diversity in subset. Our approach effectively balances these two factors, enabling the identification of high-quality samples that contribute to enhancing model performance.

K-Center \cite{sener2018active} and DQ \cite{zhou2023dataset} focus on maximizing diversity within the subset. While these approaches increase inter-sample diversity, they do not guarantee alignment with the original dataset's distribution. Both DQ and K-Center achieve 64.6\% on HumanEval, which is lower than our method. This suggests that while diversity is important, it must be balanced with distribution consistency to fully leverage the dataset's structure.
PPL and IFD prioritize the selection of complex samples, aiming to enhance model performance by focusing on challenging instances. 
However, these methods perform poorly compared to random sampling on several benchmarks. For instance, IFD shows a substantial performance drop on MBPP and MBPP+, indicating that complexity-focused sampling may deviate from the original distribution.
DEITA combines complexity and quality assessment, yielding competitive results. However, it requires the training of two additional scoring models, resulting in higher computational costs and longer training times. In contrast, our method achieves superior performance with significantly lower computational overhead, making it more efficient and scalable.

In summary, our method establishes distribution consistency as the foundation while incorporating diversity constraints, constructing a more comprehensive and efficient data selection strategy. This approach enables the identification of high-quality training samples and consistently improves the model's overall performance.

\subsection{Sampling Efficiency}

We also compared the sampling efficiency of different methods, with the results presented in Table~\ref{tab:sampling_methods}. The sampling process can be divided into two main phases: (1) the processing phase, which involves data preprocessing steps such as feature extraction and model scoring, and (2) the sampling phase, where specific strategies are applied for sample selection.

Among all compared methods, DQ exhibits the lowest sampling efficiency. Its processing phase requires nearly 20 hours to partition the dataset into non-overlapping bins, a duration that even exceeds the model training time. Although DQ achieves competitive performance, its high time cost during the sampling process represents a significant disadvantage. 
In contrast, the DEITA, PPL, and IFD methods show better efficiency but still require forward inference for each data sample. Specifically, DEITA relies on two 13B-parameter scoring models, while PPL and IFD depend on the model being trained for scoring. As a result, their sampling time increases linearly with the size of the dataset, resulting in poor scalability. 
The K-Center method reduces sampling time to 42.5 minutes by iteratively computing the similarity between candidate samples and the selected subset. However, its reliance on distance-based computations fundamentally limits further efficiency improvements.
In contrast, our method achieves the highest sampling efficiency, requiring only 13.5 minutes. This advantage is attributed to the learnable parametric model for data selection, which significantly reduces computational overhead by optimizing the loss function in Equation~\ref{eq:7} that ensures both distribution consistency with the original dataset and diversity within the selected subset.

In summary, our method maintains model performance while significantly improving sampling efficiency and scalability. Compared to other methods, our approach minimizes sampling time, enabling the efficient processing of large-scale datasets.

\begin{table*}[t]
\centering
\begin{tabular}{lcccccc}
\hline
\textbf{Model} & \textbf{Params} & \textbf{Data Size} & \textbf{HumanEval} & \textbf{HumanEval+} & \textbf{MBPP} & \textbf{MBPP+} \\
\hline
CodeLlama-Python & 7B & - & 37.8\% & 35.4\% & 59.8\% & 46.8\% \\
WizardCoder-CL & 7B & 78K & 50.6\% & 45.1\% & 58.5\% & 49.5\% \\
DeepseekCoder-Base & 6.7B & - & 47.6\% & 39.6\% & 72.0\% & 58.7\% \\
WaveCoder-DS & 6.7B & 20K & 61.0\% & 54.9\% & 75.9\% & 60.9\% \\
Magicoder-DS & 6.7B & 75K & 66.5\% & 60.4\% & 75.4\% & 61.9\% \\
Ours-DS & 6.7B & 10K & \textbf{69.5\%} & \textbf{63.4\%} & \textbf{77.2\%} & \textbf{63.2\%} \\
\hline
\end{tabular}
\caption{Comparison of different code LLMs. The results of Magicoder-DS are taken from their original paper \cite{wei2024magicoder}. We re-evaluated WaveCoder-DS as the results in their original paper \cite{yu2024wavecoder} were incomplete. Results of other models are sourced from the EvalPlus Leaderboard \cite{liu2023your}.}
\label{tab:different_llms}
\end{table*}

\subsection{Sampling Quantity}

To investigate the impact of sampling quantity on model performance, we compared the experimental results with 5K, 10K, 15K, 20K, 25K, and the full 92K dataset, as shown in Figure~\ref{fig:sampling_quantity}. The experimental results reveal that the performance curves of all methods generally exhibit an initial increase followed by a decline, indicating the presence of low-quality data within the dataset. Moderate sampling (e.g., 10K data) effectively filters out low-quality samples, thus improving model performance, while excessive sampling (more than 25K data) may introduce noise, leading to performance degradation. When trained on the full dataset, the model achieved 67.1\% on HumanEval and 74.9\% on MBPP.

The experimental results demonstrate that our method consistently outperforms others across different sampling quantities, particularly peaking at 10K samples with 69.5\% on HumanEval and 77.2\% on MBPP. Compared to training on the full dataset, our method represents improvements of 2.4\% and 2.3\% respectively. These results validate the effectiveness of our data selection strategy, showing that introducing the parametric model into the data selection process can significantly enhance model performance.

As the sampling quantity increases from 5K to 20K, PPL and IFD show performance improvements on both HumanEval and MBPP benchmarks, indicating that ``non-complex yet important'' data samples also play a key role in improving performance. The improvement observed in these methods suggests that strategies focused solely on complex data sampling fail to yield the best performance, as they neglect simpler yet impactful samples. In contrast, our method considers both distribution consistency and diversity, ensuring that the sampled dataset not only includes complex data but also retains simpler yet essential samples, resulting in a more balanced performance gains.

The performance curves of DEITA, DQ and K-Center exhibit similar trends, with performance declining after exceeding 15K samples. Notably, K-Center maintains relatively good performance on the MBPP benchmark but fluctuates significantly on the HumanEval benchmark. This reflects the limitations of diversity-driven sampling strategies in programming semantic understanding. Purely diversity-based approaches neglect distribution consistency, potentially resulting in unstable performance on different benchmarks.

Overall, our method achieves stable and superior performance across different sampling quantities by balancing both distribution consistency and diversity. In particular, with a 10K sample size, the model reaches optimal performance. This demonstrates that our approach not only excels in improving performance but also effectively avoids the noise introduced by excessive sampling, thus enhancing the model's generalization ability. Our data selection strategy enables the identification of high-quality samples, significantly boosting the model's overall performance.

\subsection{Code LLMs}

We compared the performance of different code LLMs, as shown in Table~\ref{tab:different_llms}. WizardCoder-CL was trained on CodeLlama-Python using 78K instruction samples, while WaveCoder-DS and Magicoder-DS were both trained on DeepSeekCoder-Base with 20K and 75K samples, respectively. Despite these models utilizing much larger training datasets, our approach achieves superior performance across all benchmarks using only 10K carefully selected samples.

These results highlight that data quality is more important than quantity. By developing an effective data selection strategy, we can significantly enhance model performance. Our experiments confirm that employing the parametric model for data selection constitutes an efficient approach. It not only identifies high-quality training samples but also substantially improves training efficiency by reducing the number of unnecessary or low-quality samples. Compared to models relying on massive training data, our method delivers better performance with far fewer data, demonstrating that our data selection strategy ensures competitive performance while effectively reducing computational resource consumption.

\section{Conclusion}

In this work, we propose an efficient data selection strategy designed for code data. By optimizing parametric models, our approach ensures both distribution consistency between the selected subset and the original dataset, while simultaneously maximizing the diversity of the subset. Experimental results demonstrate that by using only 10K carefully selected data, our method achieves the best performance across all benchmarks, outperforming both other sampling methods and existing code LLMs. These findings underscore that the data quality is more important than the data quantity, and that an effective data selection strategies can significantly enhance both model performance and training efficiency.
We hope that our study provides valuable insights for efficiently training code LLMs and contributes to advancing progress in related research fields.

\section*{Limitations}

While our method shows significant advantages in code generation tasks, there are limitations that warrant further investigation.
First, the current experimental validation is limited to Python-specific evaluation benchmarks and is tested solely on the DeepSeekCoder-Base-6.7B model. Due to expensive equipment rental costs, we have not yet conducted experiments on other programming languages (e.g., Java, C++) or larger-scale models (e.g., 34B parameters).
More critically, our method does not directly validate the functional correctness of the code, potentially allowing flawed samples into the training set. 
In future work, we plan to integrate executable sandbox environments to rigorously verify code correctness and extend evaluations to a broader range of programming languages and model scales for a comprehensive assessment of the method's generalization capabilities.

\section*{Acknowledgments}

We would like to express our sincere gratitude to those who have contributed to this work. Additionally, we utilize ChatGPT to review and refine sentence structures, which helped enhance the clarity and quality of the text.

\bibliography{ref}

\begin{thebibliography}{41}
\providecommand{\natexlab}[1]{#1}

\bibitem[{Achiam et~al.(2023)Achiam, Adler, Agarwal, Ahmad, Akkaya, Aleman, Almeida, Altenschmidt, Altman, Anadkat et~al.}]{achiam2023gpt}
Josh Achiam, Steven Adler, Sandhini Agarwal, Lama Ahmad, Ilge Akkaya, Florencia~Leoni Aleman, Diogo Almeida, Janko Altenschmidt, Sam Altman, Shyamal Anadkat, and 1 others. 2023.
\newblock Gpt-4 technical report.
\newblock \emph{arXiv preprint arXiv:2303.08774}.

\bibitem[{Ankner et~al.(2024)Ankner, Blakeney, Sreenivasan, Marion, Leavitt, and Paul}]{ankner2024perplexed}
Zachary Ankner, Cody Blakeney, Kartik Sreenivasan, Max Marion, Matthew~L Leavitt, and Mansheej Paul. 2024.
\newblock Perplexed by perplexity: Perplexity-based data pruning with small reference models.
\newblock \emph{arXiv preprint arXiv:2405.20541}.

\bibitem[{Austin et~al.(2021)Austin, Odena, Nye, Bosma, Michalewski, Dohan, Jiang, Cai, Terry, Le et~al.}]{austin2021program}
Jacob Austin, Augustus Odena, Maxwell Nye, Maarten Bosma, Henryk Michalewski, David Dohan, Ellen Jiang, Carrie Cai, Michael Terry, Quoc Le, and 1 others. 2021.
\newblock Program synthesis with large language models.
\newblock \emph{arXiv preprint arXiv:2108.07732}.

\bibitem[{Chaudhary(2023)}]{codealpaca}
Sahil Chaudhary. 2023.
\newblock Code alpaca: An instruction-following llama model for code generation.
\newblock \url{https://github.com/sahil280114/codealpaca}.

\bibitem[{Chen et~al.(2024)Chen, Li, Yan, Wang, Gunaratna, Yadav, Tang, Srinivasan, Zhou, Huang et~al.}]{chen2024alpagasus}
Lichang Chen, Shiyang Li, Jun Yan, Hai Wang, Kalpa Gunaratna, Vikas Yadav, Zheng Tang, Vijay Srinivasan, Tianyi Zhou, Heng Huang, and 1 others. 2024.
\newblock Alpagasus: Training a better alpaca with fewer data.
\newblock In \emph{The Twelfth International Conference on Learning Representations}.

\bibitem[{Chen et~al.(2021)Chen, Tworek, Jun, Yuan, Pinto, Kaplan, Edwards, Burda, Joseph, Brockman et~al.}]{chen2021evaluating}
Mark Chen, Jerry Tworek, Heewoo Jun, Qiming Yuan, Henrique Ponde De~Oliveira Pinto, Jared Kaplan, Harri Edwards, Yuri Burda, Nicholas Joseph, Greg Brockman, and 1 others. 2021.
\newblock Evaluating large language models trained on code.
\newblock \emph{arXiv preprint arXiv:2107.03374}.

\bibitem[{Chung et~al.(2024)Chung, Hou, Longpre, Zoph, Tay, Fedus, Li, Wang, Dehghani, Brahma et~al.}]{chung2024scaling}
Hyung~Won Chung, Le~Hou, Shayne Longpre, Barret Zoph, Yi~Tay, William Fedus, Yunxuan Li, Xuezhi Wang, Mostafa Dehghani, Siddhartha Brahma, and 1 others. 2024.
\newblock Scaling instruction-finetuned language models.
\newblock \emph{Journal of Machine Learning Research}, 25(70):1--53.

\bibitem[{Gunasekar et~al.(2023)Gunasekar, Zhang, Aneja, Mendes, Del~Giorno, Gopi, Javaheripi, Kauffmann, de~Rosa, Saarikivi et~al.}]{gunasekar2023textbooks}
Suriya Gunasekar, Yi~Zhang, Jyoti Aneja, Caio C{\'e}sar~Teodoro Mendes, Allie Del~Giorno, Sivakanth Gopi, Mojan Javaheripi, Piero Kauffmann, Gustavo de~Rosa, Olli Saarikivi, and 1 others. 2023.
\newblock Textbooks are all you need.
\newblock \emph{arXiv preprint arXiv:2306.11644}.

\bibitem[{Guo et~al.(2025)Guo, Yang, Zhang, Song, Zhang, Xu, Zhu, Ma, Wang, Bi et~al.}]{guo2025deepseek}
Daya Guo, Dejian Yang, Haowei Zhang, Junxiao Song, Ruoyu Zhang, Runxin Xu, Qihao Zhu, Shirong Ma, Peiyi Wang, Xiao Bi, and 1 others. 2025.
\newblock Deepseek-r1: Incentivizing reasoning capability in llms via reinforcement learning.
\newblock \emph{arXiv preprint arXiv:2501.12948}.

\bibitem[{Guo et~al.(2024)Guo, Zhu, Yang, Xie, Dong, Zhang, Chen, Bi, Wu, Li et~al.}]{guo2024deepseek}
Daya Guo, Qihao Zhu, Dejian Yang, Zhenda Xie, Kai Dong, Wentao Zhang, Guanting Chen, Xiao Bi, Yu~Wu, YK~Li, and 1 others. 2024.
\newblock Deepseek-coder: When the large language model meets programming--the rise of code intelligence.
\newblock \emph{arXiv preprint arXiv:2401.14196}.

\bibitem[{Hui et~al.(2024)Hui, Yang, Cui, Yang, Liu, Zhang, Liu, Zhang, Yu, Lu et~al.}]{hui2024qwen2}
Binyuan Hui, Jian Yang, Zeyu Cui, Jiaxi Yang, Dayiheng Liu, Lei Zhang, Tianyu Liu, Jiajun Zhang, Bowen Yu, Keming Lu, and 1 others. 2024.
\newblock Qwen2.5-coder technical report.
\newblock \emph{arXiv preprint arXiv:2409.12186}.

\bibitem[{Hurst et~al.(2024)Hurst, Lerer, Goucher, Perelman, Ramesh, Clark, Ostrow, Welihinda, Hayes, Radford et~al.}]{hurst2024gpt}
Aaron Hurst, Adam Lerer, Adam~P Goucher, Adam Perelman, Aditya Ramesh, Aidan Clark, AJ~Ostrow, Akila Welihinda, Alan Hayes, Alec Radford, and 1 others. 2024.
\newblock Gpt-4o system card.
\newblock \emph{arXiv preprint arXiv:2410.21276}.

\bibitem[{Iyer et~al.(2021)Iyer, Khargoankar, Bilmes, and Asanani}]{iyer2021submodular}
Rishabh Iyer, Ninad Khargoankar, Jeff Bilmes, and Himanshu Asanani. 2021.
\newblock Submodular combinatorial information measures with applications in machine learning.
\newblock In \emph{Algorithmic Learning Theory}, pages 722--754. PMLR.

\bibitem[{Li et~al.(2023{\natexlab{a}})Li, Hammoud, Itani, Khizbullin, and Ghanem}]{li2023camel}
Guohao Li, Hasan Hammoud, Hani Itani, Dmitrii Khizbullin, and Bernard Ghanem. 2023{\natexlab{a}}.
\newblock Camel: Communicative agents for" mind" exploration of large language model society.
\newblock \emph{Advances in Neural Information Processing Systems}, 36:51991--52008.

\bibitem[{Li et~al.(2024{\natexlab{a}})Li, Zhang, He, Li, Zhao, Wang, Cheng, and Zhou}]{li2024superfiltering}
Ming Li, Yong Zhang, Shwai He, Zhitao Li, Hongyu Zhao, Jianzong Wang, Ning Cheng, and Tianyi Zhou. 2024{\natexlab{a}}.
\newblock Superfiltering: Weak-to-strong data filtering for fast instruction-tuning.
\newblock In \emph{Proceedings of the 62nd Annual Meeting of the Association for Computational Linguistics (Volume 1: Long Papers)}, pages 14255--14273.

\bibitem[{Li et~al.(2024{\natexlab{b}})Li, Zhang, Li, Chen, Chen, Cheng, Wang, Zhou, and Xiao}]{li2024quantity}
Ming Li, Yong Zhang, Zhitao Li, Jiuhai Chen, Lichang Chen, Ning Cheng, Jianzong Wang, Tianyi Zhou, and Jing Xiao. 2024{\natexlab{b}}.
\newblock From quantity to quality: Boosting llm performance with self-guided data selection for instruction tuning.
\newblock In \emph{Proceedings of the 2024 Conference of the North American Chapter of the Association for Computational Linguistics: Human Language Technologies (Volume 1: Long Papers)}, pages 7595--7628.

\bibitem[{Li et~al.(2023{\natexlab{b}})Li, Allal, Zi, Muennighoff, Kocetkov, Mou, Marone, Akiki, Li, Chim et~al.}]{li2023starcoder}
Raymond Li, Loubna~Ben Allal, Yangtian Zi, Niklas Muennighoff, Denis Kocetkov, Chenghao Mou, Marc Marone, Christopher Akiki, Jia Li, Jenny Chim, and 1 others. 2023{\natexlab{b}}.
\newblock Starcoder: may the source be with you!
\newblock \emph{arXiv preprint arXiv:2305.06161}.

\bibitem[{Liu et~al.(2024{\natexlab{a}})Liu, Feng, Xue, Wang, Wu, Lu, Zhao, Deng, Zhang, Ruan et~al.}]{liu2024deepseek}
Aixin Liu, Bei Feng, Bing Xue, Bingxuan Wang, Bochao Wu, Chengda Lu, Chenggang Zhao, Chengqi Deng, Chenyu Zhang, Chong Ruan, and 1 others. 2024{\natexlab{a}}.
\newblock Deepseek-v3 technical report.
\newblock \emph{arXiv preprint arXiv:2412.19437}.

\bibitem[{Liu et~al.(2023)Liu, Xia, Wang, and Zhang}]{liu2023your}
Jiawei Liu, Chunqiu~Steven Xia, Yuyao Wang, and Lingming Zhang. 2023.
\newblock Is your code generated by chatgpt really correct? rigorous evaluation of large language models for code generation.
\newblock \emph{Advances in Neural Information Processing Systems}, 36:21558--21572.

\bibitem[{Liu et~al.(2024{\natexlab{b}})Liu, Zeng, He, Jiang, and He}]{liu2024makes}
Wei Liu, Weihao Zeng, Keqing He, Yong Jiang, and Junxian He. 2024{\natexlab{b}}.
\newblock What makes good data for alignment? a comprehensive study of automatic data selection in instruction tuning.
\newblock In \emph{The Twelfth International Conference on Learning Representations}.

\bibitem[{Lozhkov et~al.(2024)Lozhkov, Li, Allal, Cassano, Lamy-Poirier, Tazi, Tang, Pykhtar, Liu, Wei et~al.}]{lozhkov2024starcoder}
Anton Lozhkov, Raymond Li, Loubna~Ben Allal, Federico Cassano, Joel Lamy-Poirier, Nouamane Tazi, Ao~Tang, Dmytro Pykhtar, Jiawei Liu, Yuxiang Wei, and 1 others. 2024.
\newblock Starcoder 2 and the stack v2: The next generation.
\newblock \emph{arXiv preprint arXiv:2402.19173}.

\bibitem[{Lu et~al.(2024)Lu, Yuan, Yuan, Lin, Lin, Tan, Zhou, and Zhou}]{lu2024instag}
Keming Lu, Hongyi Yuan, Zheng Yuan, Runji Lin, Junyang Lin, Chuanqi Tan, Chang Zhou, and Jingren Zhou. 2024.
\newblock \# instag: Instruction tagging for analyzing supervised fine-tuning of large language models.
\newblock In \emph{The Twelfth International Conference on Learning Representations}.

\bibitem[{Luo et~al.(2023)Luo, Xu, Zhao, Sun, Geng, Hu, Tao, Ma, Lin, and Jiang}]{luowizardcoder}
Ziyang Luo, Can Xu, Pu~Zhao, Qingfeng Sun, Xiubo Geng, Wenxiang Hu, Chongyang Tao, Jing Ma, Qingwei Lin, and Daxin Jiang. 2023.
\newblock Wizardcoder: Empowering code large language models with evol-instruct.
\newblock In \emph{The Twelfth International Conference on Learning Representations}.

\bibitem[{Lv et~al.(2025)Lv, Xia, and Huang}]{lv2025data}
Weijie Lv, Xuan Xia, and Sheng-Jun Huang. 2025.
\newblock Data-efficient llm fine-tuning for code generation.
\newblock \emph{arXiv preprint arXiv:2504.12687}.

\bibitem[{Reimers and Gurevych(2019)}]{reimers-2019-sentence-bert}
Nils Reimers and Iryna Gurevych. 2019.
\newblock \href {https://arxiv.org/abs/1908.10084} {Sentence-bert: Sentence embeddings using siamese bert-networks}.
\newblock In \emph{Proceedings of the 2019 Conference on Empirical Methods in Natural Language Processing}. Association for Computational Linguistics.

\bibitem[{Roziere et~al.(2023)Roziere, Gehring, Gloeckle, Sootla, Gat, Tan, Adi, Liu, Sauvestre, Remez et~al.}]{roziere2023code}
Baptiste Roziere, Jonas Gehring, Fabian Gloeckle, Sten Sootla, Itai Gat, Xiaoqing~Ellen Tan, Yossi Adi, Jingyu Liu, Romain Sauvestre, Tal Remez, and 1 others. 2023.
\newblock Code llama: Open foundation models for code.
\newblock \emph{arXiv preprint arXiv:2308.12950}.

\bibitem[{Sener and Savarese(2018)}]{sener2018active}
Ozan Sener and Silvio Savarese. 2018.
\newblock Active learning for convolutional neural networks: A core-set approach.
\newblock In \emph{International Conference on Learning Representations}.

\bibitem[{Team et~al.(2024)Team, Zhao, Hui, Howland, Nguyen, Zuo, Hu, Choquette-Choo, Shen, Kelley et~al.}]{team2024codegemma}
CodeGemma Team, Heri Zhao, Jeffrey Hui, Joshua Howland, Nam Nguyen, Siqi Zuo, Andrea Hu, Christopher~A Choquette-Choo, Jingyue Shen, Joe Kelley, and 1 others. 2024.
\newblock Codegemma: Open code models based on gemma.
\newblock \emph{arXiv preprint arXiv:2406.11409}.

\bibitem[{Touvron et~al.(2023)Touvron, Martin, Stone, Albert, Almahairi, Babaei, Bashlykov, Batra, Bhargava, Bhosale et~al.}]{touvron2023llama}
Hugo Touvron, Louis Martin, Kevin Stone, Peter Albert, Amjad Almahairi, Yasmine Babaei, Nikolay Bashlykov, Soumya Batra, Prajjwal Bhargava, Shruti Bhosale, and 1 others. 2023.
\newblock Llama 2: Open foundation and fine-tuned chat models.
\newblock \emph{arXiv preprint arXiv:2307.09288}.

\bibitem[{Wang et~al.(2024)Wang, Liu, Shi, Li, Chen, Lu, and Yang}]{wang2024inscl}
Yifan Wang, Yafei Liu, Chufan Shi, Haoling Li, Chen Chen, Haonan Lu, and Yujiu Yang. 2024.
\newblock Inscl: A data-efficient continual learning paradigm for fine-tuning large language models with instructions.
\newblock In \emph{Proceedings of the 2024 Conference of the North American Chapter of the Association for Computational Linguistics: Human Language Technologies (Volume 1: Long Papers)}, pages 663--677.

\bibitem[{Wang et~al.(2023)Wang, Kordi, Mishra, Liu, Smith, Khashabi, and Hajishirzi}]{wang2023self}
Yizhong Wang, Yeganeh Kordi, Swaroop Mishra, Alisa Liu, Noah~A Smith, Daniel Khashabi, and Hannaneh Hajishirzi. 2023.
\newblock Self-instruct: Aligning language models with self-generated instructions.
\newblock In \emph{Proceedings of the 61st Annual Meeting of the Association for Computational Linguistics (Volume 1: Long Papers)}, pages 13484--13508.

\bibitem[{Wei et~al.(2021)Wei, Bosma, Zhao, Guu, Yu, Lester, Du, Dai, and Le}]{wei2021finetuned}
Jason Wei, Maarten Bosma, Vincent~Y Zhao, Kelvin Guu, Adams~Wei Yu, Brian Lester, Nan Du, Andrew~M Dai, and Quoc~V Le. 2021.
\newblock Finetuned language models are zero-shot learners.
\newblock \emph{arXiv preprint arXiv:2109.01652}.

\bibitem[{Wei et~al.(2024)Wei, Wang, Liu, Ding, and Zhang}]{wei2024magicoder}
Yuxiang Wei, Zhe Wang, Jiawei Liu, Yifeng Ding, and Lingming Zhang. 2024.
\newblock Magicoder: Empowering code generation with oss-instruct.
\newblock In \emph{International Conference on Machine Learning}, pages 52632--52657. PMLR.

\bibitem[{Xia et~al.(2024)Xia, Malladi, Gururangan, Arora, and Chen}]{xia2024less}
Mengzhou Xia, Sadhika Malladi, Suchin Gururangan, Sanjeev Arora, and Danqi Chen. 2024.
\newblock Less: selecting influential data for targeted instruction tuning.
\newblock In \emph{Proceedings of the 41st International Conference on Machine Learning}, pages 54104--54132.

\bibitem[{Xie et~al.(2023)Xie, Lu, Yan, Yang, Tomizuka, and Zhan}]{xie2023active}
Yichen Xie, Han Lu, Junchi Yan, Xiaokang Yang, Masayoshi Tomizuka, and Wei Zhan. 2023.
\newblock Active finetuning: Exploiting annotation budget in the pretraining-finetuning paradigm.
\newblock In \emph{Proceedings of the IEEE/CVF Conference on Computer Vision and Pattern Recognition}, pages 23715--23724.

\bibitem[{Xu et~al.(2024)Xu, Sun, Zheng, Geng, Zhao, Feng, Tao, Lin, and Jiang}]{xu2024wizardlm}
Can Xu, Qingfeng Sun, Kai Zheng, Xiubo Geng, Pu~Zhao, Jiazhan Feng, Chongyang Tao, Qingwei Lin, and Daxin Jiang. 2024.
\newblock \href {https://openreview.net/forum?id=CfXh93NDgH} {Wizardlm: Empowering large pre-trained language models to follow complex instructions}.
\newblock In \emph{The Twelfth International Conference on Learning Representations}.

\bibitem[{Xu et~al.(2023)Xu, Yao, Huang, Qi, Wang, Gu, and Sundaresan}]{xu2023rethinking}
Yang Xu, Yongqiang Yao, Yufan Huang, Mengnan Qi, Maoquan Wang, Bin Gu, and Neel Sundaresan. 2023.
\newblock Rethinking the instruction quality: Lift is what you need.
\newblock \emph{arXiv preprint arXiv:2312.11508}.

\bibitem[{Yang et~al.(2024)Yang, Yang, Zhang, Hui, Zheng, Yu, Li, Liu, Huang, Wei et~al.}]{yang2024qwen2}
An~Yang, Baosong Yang, Beichen Zhang, Binyuan Hui, Bo~Zheng, Bowen Yu, Chengyuan Li, Dayiheng Liu, Fei Huang, Haoran Wei, and 1 others. 2024.
\newblock Qwen2.5 technical report.
\newblock \emph{arXiv preprint arXiv:2412.15115}.

\bibitem[{Yu et~al.(2024)Yu, Zhang, Shang, Huang, Xu, Zhao, Hu, and Yin}]{yu2024wavecoder}
Zhaojian Yu, Xin Zhang, Ning Shang, Yangyu Huang, Can Xu, Yishujie Zhao, Wenxiang Hu, and Qiufeng Yin. 2024.
\newblock Wavecoder: Widespread and versatile enhancement for code large language models by instruction tuning.
\newblock In \emph{Proceedings of the 62nd Annual Meeting of the Association for Computational Linguistics (Volume 1: Long Papers)}, pages 5140--5153.

\bibitem[{Zhou et~al.(2023{\natexlab{a}})Zhou, Liu, Xu, Iyer, Sun, Mao, Ma, Efrat, Yu, Yu et~al.}]{zhou2023lima}
Chunting Zhou, Pengfei Liu, Puxin Xu, Srinivasan Iyer, Jiao Sun, Yuning Mao, Xuezhe Ma, Avia Efrat, Ping Yu, Lili Yu, and 1 others. 2023{\natexlab{a}}.
\newblock Lima: Less is more for alignment.
\newblock \emph{Advances in Neural Information Processing Systems}, 36:55006--55021.

\bibitem[{Zhou et~al.(2023{\natexlab{b}})Zhou, Wang, Gu, Peng, Lian, Zhang, You, and Feng}]{zhou2023dataset}
Daquan Zhou, Kai Wang, Jianyang Gu, Xiangyu Peng, Dongze Lian, Yifan Zhang, Yang You, and Jiashi Feng. 2023{\natexlab{b}}.
\newblock Dataset quantization.
\newblock In \emph{Proceedings of the IEEE/CVF International Conference on Computer Vision}, pages 17205--17216.

\end{thebibliography}

\appendix



\end{document}